\documentclass{article}
\usepackage{spconf,amsmath,graphicx}
\usepackage{pbox}
\usepackage{times}
\usepackage{latexsym}
\usepackage{color}
\usepackage{times}
\usepackage{bbm}
\usepackage{dsfont}
\usepackage{qtree}
\usepackage{multirow}
\usepackage{url}
\usepackage{latexsym}
\usepackage{graphicx}
\usepackage{amsmath,amssymb,amsthm}

\usepackage{url}
\usepackage{diagbox}
\usepackage{fancyvrb}
\usepackage{tabularx}

\title{Nudging Neural Conversational Model with Domain Knowledge}
\name{Sungjin Lee}
\address{Microsoft Research, Redmond, WA, USA}
\begin{document}
%
\maketitle
\begin{abstract}
Neural conversation models are attractive because one can train a model directly on dialog examples with minimal labeling. With a small amount of data, however, they often fail to generalize over test data since they tend to capture spurious features instead of semantically meaningful domain knowledge.
To address this issue, we propose a novel approach that allows any human teachers to transfer their domain knowledge to the conversation model in the form of natural language rules. 
We tested our method with three different dialog datasets. The improved performance across all domains demonstrates the efficacy of our proposed method.
\end{abstract}

\section{Introduction}
\label{sec:intro}

Recently, conversational systems have been increasingly adopting neural approaches~\cite{sordoni2015neural,vinyals2015neural,serban2016building,wen2016network,bordes2016learning,williams2017hybrid,zhao2017generative}. Neural approaches are attractive because they allow us to directly train a model on dialog examples with minimal labeling, which significantly reduces the development complexity compared to traditional approaches~\cite{jokinen2009spoken,young2013pomdp}. Now neural networks are at the center of services like Conversation Learner~\footnote{https://labs.cognitive.microsoft.com/en-us/project-conversation-learner} and Rasa~\footnote{http://rasa.com/} which allow developers to interactively build bots with much less hand-crafted features.
For task-oriented conversational systems, however, neural approaches still have many challenges to overcome. 
Particularly, the overfitting problem of neural approaches can be severe when there is an insufficient amount of dialog examples available. It is mainly because the model over-optimize on training data by capturing spurious features instead of learning actually meaningful features for the task. 

\begin{table}[t]
\centering
\small
\begin{tabular}{l}
\hline\hline
... \\ System: I'm on it \\ User: Actually I would prefer for two people \\ System: Sure, is there anything else to update? \\ \hline
... \\ System: I'm on it \\ User: Actually I want French food \\ System: Sure, is there anything else to update? \\ \hline\hline
\end{tabular}
\caption{Example dialogs in a restaurant finding domain}
\label{tab:examples}
\end{table}

In this work, we address this problem with a new paradigm, so-called, {\em machine teaching}~\cite{simard2017machine}. In machine teaching, the role of the teacher is to transfer knowledge to the learning machine so that it can generate a useful model. With their domain knowledge, the teachers can divine some features that are immune to overfitting because they are created independently of the training data. In Table~\ref{tab:examples}, for example, the learning machine might pick the previous system response, i.e. \texttt{I'm on it}, as a key feature in making predictions on next response because it consistently shows up in the limited set of dialogs. But such spurious regularities won't last long as more dialogs become available. In contrast, any human teacher can tell that the user inputs should instead be the key feature despite their varying surface forms. Thus, if there is an easy way to transfer such knowledge, one can bias the model to rely more on semantically robust features.

\begin{figure*}[!t]
  \begin{center}
    \includegraphics[width=0.8\textwidth]{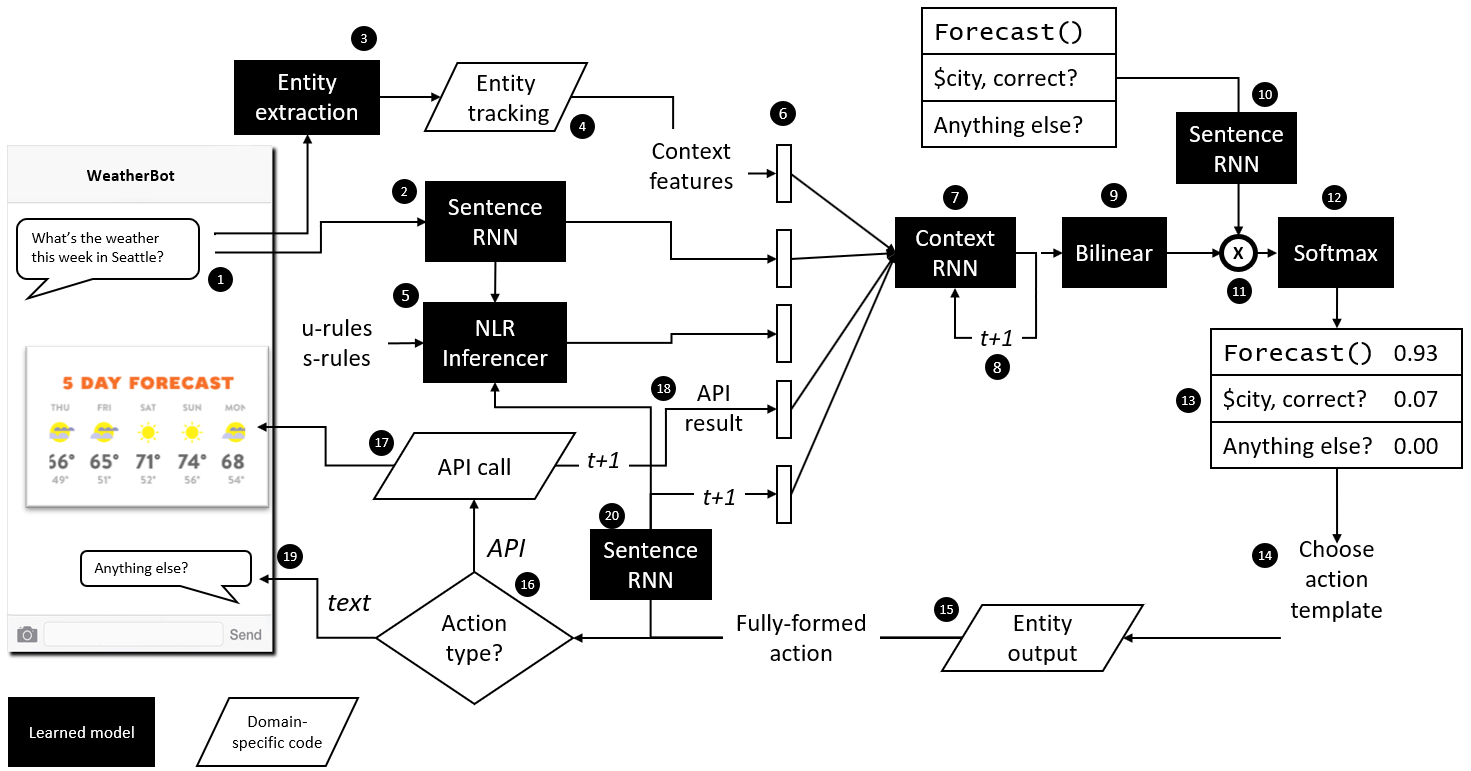}
  \end{center}
  \caption{\small The overall operational loop. Trapezoids refer to programmatic code provided by the software developer, and shaded boxes are trainable components.}
  \label{fig:overall}
\end{figure*}

As a first step toward this goal, we propose a novel conversational model which allows one to express domain knowledge in the form of {\em Natural Language Rules} (NLRs).~\footnote{We think the use of natural language for knowledge description is important for a wide adoption because representing knowledge with formal languages is a non-trivial skill to master.} Specifically, our method takes as input two sets of rules, {\em u-rules} and {\em s-rules}. With {\em u-rules}, one can suggest what the system should say upon a particular user input, e.g, `Actually I want \$cusine food' $\rightarrow$ `Sure, is there anything else to update?'.~\footnote{In this example, we delexicalized the utterance by substituting the actual value (e.g. French) with its category.} 
Whereas, with {\em s-rules}, one can encourage the system to follow a typical ordering of system actions, e.g., `what kind of food would you like?' $\rightarrow$ `what area of town should I search?'. 
With the {\em NLR inferencer} (introduced in Section~\ref{sec:nlr}), we perform inference based on NLRs. 
The inference result is passed to the overall conversational model as features to give it a nudge to consider domain knowledge. 

The rest of this paper is organized as follows. In Section~\ref{sec:method} we describe our conversational model with an NLR inference module. In Section~\ref{sec:experiments} we discuss our experiments and results. We finish with conclusions and future work in Section~\ref{sec:conclusion}.

\section{Conversation Model with Natural Language Rules}
\label{sec:method}
In this section, we describe our task-oriented conversation model that makes use of domain-specific rules that are expressed in natural language.

\subsection{Overall Architecture}
\label{overall}
\noindent
\begin{figure*}[!t]
  \begin{center}
    \includegraphics[width=0.7\textwidth]{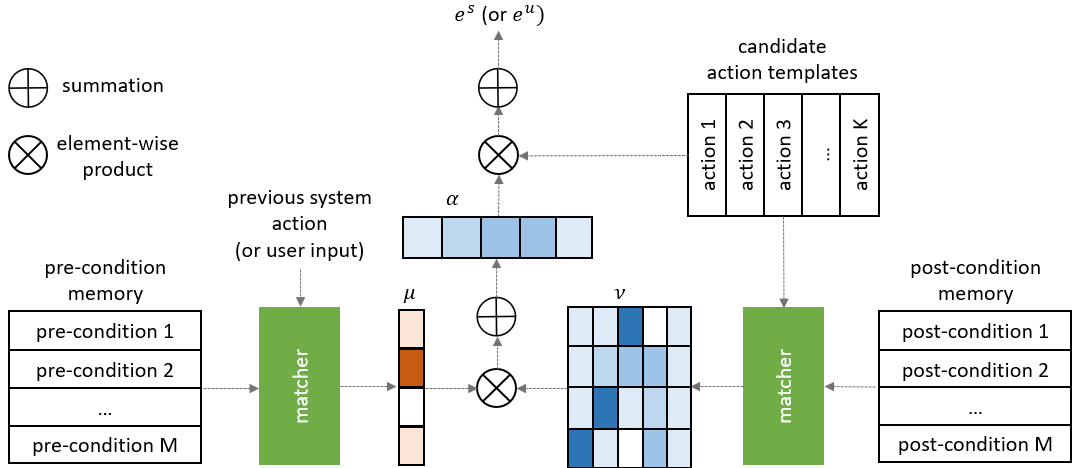}
  \end{center}
  \caption{\small Natural Language Rule-based inference process}
  \label{fig:nlr}
\end{figure*}
Our conversational model builds upon the Hybrid Code Network (HCN)~\cite{williams2017hybrid} which servers as a key component for the recent interactive bot development technologies such as Conversation Learner and Rasa.  
At a high level, our model have five components: sentence RNNs; a context RNN; domain-specific software; domain-specific action templates; and a conventional entity extraction module for identifying entity mentions in text. We use a pre-trained sentence encoder obtained from~\cite{li2016persona} which encodes a sentence with 4 layer LSTM models~\cite{hochreiter1997long} with 1,000 hidden units for each layer. We take the hidden state from the top layer at the end of the sentence as the sentence representation. We call this sentence encoder {\em Neurocon} and use it for all sentence RNNs.
Both the context RNN and the developer code maintain state. Each action template can be a textual communicative action or an API call. 
Figure~\ref{fig:overall} shows the overall operational loop.
The cycle begins when the user provides an utterance, as text (step 1). Second, an utterance embedding is formed, using the sentence RNN (step 2). Third, an entity extraction module identifies entity mentions (step 3), for example, identifying ``Seattle'' as a \texttt{$\langle$city$\rangle$} entity. The text and entity mentions are then passed to ``Entity tracking'' code provided by the developer (step 4), which grounds and maintains entities, for example, mapping the text ``Seattle'' to a specific row in a database. This code can optionally return ``context features'' which are features the developer thinks will be useful for distinguishing among actions, such as which entities are currently present and which are absent. An embedding for the previous system's action is generated (step 5). Then, we feed the NLR inferencer with the user utterance embedding, system action embedding and the two sets of NLRs, {\em u-rules} and {\em s-rules}. The NLR inferencer performs inference to yield a vector that represents a soft preference over the set of action templates based on the domain knowledge encoded in the NLRs (step 6). 
The feature components from steps 1-6 are concatenated to form a feature vector (step 7). This vector is passed to the context RNN. The context RNN computes a hidden state (step 8), which is retained for the next timestep (step 9), and passed to a bilinear layer which projects the hidden state to the response space (step 10), yielding a response embedding. A set of embeddings for each distinct system action template are generated, using the sentence RNN (step 11). This set of embeddings get ranked according to the similarity to the response embedding (step 12). With a softmax, a probability distribution over action templates is generated (step 13). From the resulting distribution (step 14), the best action is selected (step 15). The selected action is next passed to ``Entity output'' developer code that can substitute in entities (step 16) and produce a fully-formed action, for example, mapping the template ``$\langle$city$\rangle$, right?'' to ``Seattle, right?''. In step 17, control branches depending on the type of the action: if it is an API action, the corresponding API call in the developer code is invoked (step 18), for example, to render rich content to the user. APIs can act as sensors and return features relevant to the dialog, so these can be added to the feature vector in the next timestep (step 19). If the action is text, it is rendered to the user (step 20), and cycle then repeats. Note that there are a few improvements in our model compared to the HCN model, we encode system actions using RNNs rather than just featurizing the system action taken with a binary vector which is all zero values except for the index of the taken action; we rank a set of candidate system actions by matching them with the context rather than performing classification without looking into the actions. 

\subsection{Natural Language Rule Inferencer}
\label{sec:nlr}
The Figure~\ref{fig:nlr} depicts the entire process of NLR inference. 
There are two sets of rules {\em s-rules} and {\em u-rules}. A rule is a tuple of \textsl{(pre-condition, post-condition)} -- specifically, the pre-condition and post-condition for {\em s-rules} ({\em u-rules}) are ``previous system action'' and ``system action'' (``user input'' and ``system action''), respectively.
For each rule set, we store pre-condition embeddings and post-condition embeddings in the pre-condition memory and post-condition memory, respectively. We use the sentence RNNs (in Section~\ref{overall}) to encode pre/post-conditions.
Then, we match the inferencer input, which is either the previous system action or the current user input depending on the rule set, against the pre-condition embeddings to generate a vector of matching scores, $\mu$ (red vector in Figure~\ref{fig:nlr}). This indicates how relevant each rule is given the inferencer input. 
At the same time, we match each of the post-condition embeddings against all candidate action templates to generate a matrix of matching scores, $\nu$ (blue matrix). Each element of this matrix represents how relevant a candidate action template is when the corresponding rule gets matched.
Thus, the element-wise product of $\mu$ and $\nu$ yields a weighted matrix such that the summation of it over the rows (i.e., rules), results in a vector, $\alpha$ (blue vector), that represents how relevant a candidate action template is based on the inferencer input and the rules.
Finally, we produce output vectors $e_s$ and $e_u$ based on {\em s-rules} and {\em u-rules}, respectively by taking a weighted average over the set of candidate action template embeddings with the weight being $\alpha$.~\footnote{Due to the limited coverage of rules, it is possible that there is no entry in the pre-condition memory (or post-condition memory) matching the inferencer input (or candidate action templates). To handle such cases, we introduce two ``no match'' bias parameters for each matching process and extend the set of candidate embeddings with a zero vector. When we compute the weighted average, the zero vector gets multiplied by the probabilities assigned to the bias terms.} Then, the NLR inferencer yields the final vector by concatenating $e^s$ and $e^u$.
For the matcher, we first measure cosine similarities and normalize them to probabilities through the softmax operation: $m = \textsl{softmax}(\textsl{cosine}(a, b)/\lambda)$, where $\lambda$ is a scalar to adjust the sharpness of the resulting probability distribution $m$.

\section{Experiments}
\label{sec:experiments}
\subsection{Data}
We use three dialog datasets --- \texttt{Weather}, \texttt{Navigate}, and \texttt{Schedule}. Basic statistics of the datasets are shown in Table~\ref{tab:data_size} and Table~\ref{tab:data_stats}.
The datasets are three distinct domains of the recently released Stanford dialog data for an in-car assistant: weather information retrieval, point-of-interest navigation, and calendar scheduling, respectively~\cite{eric2017key}. Dialogs were collected through Amazon Mechanical Turk using a Wizard-of-Oz scheme.
To create datasets that are appropriate for training and testing models, we first delexicalized the datasets based on dialog act annotations and performed random negative sampling from the set of distinct action temples to generate 9 distractors for the true action template for each turn.
We created a small number of simple rules, spending time less than a half hour for each domain: 20 rules for \texttt{Weather}, 16 for \texttt{Navigate}, and 17 for \texttt{Schedule}.~\footnote{The rules are made available at \url{https://www.dropbox.com/s/qz6wgxngtzm0dzp/rules.zip?dl=0}.}
\begin{table}[h]
\centering
\small
\begin{tabular}{|c||c|c|c|c|c|c|c|}
\hline
Domain	& Train	& Dev	& Test\\ \hline
\texttt{Weather}	& 797	& 99	& 100	\\ \hline
\texttt{Navigate}	& 800	& 100	& 100	\\ \hline
\texttt{Schedule}	& 828	& 103	& 104	\\ \hline
\end{tabular}
\caption{The size of datasets}
\label{tab:data_size}
\end{table}

\begin{table}[h]
\small
\centering
\begin{tabular}{|l|c|c|c|c|}
\hline
Metric & Weather & Nav.	& Sch.	\\ \hline
Avg. turns per dialog	& 5.40	& 6.56	& 7.32	\\ \hline
Avg. tokens per user turn	& 5.51	& 7.06	& 10.06	\\ \hline
Avg. tokens per system turn	& 5.70	& 7.88	& 16.14	\\ \hline
distinct action templates	& 187	& 259	& 158	\\ \hline
\end{tabular}
\caption{Data statistics. Nav. and Sch. stand for Navigation and Schedule, respectively.}
\label{tab:data_stats}
\end{table}

\subsection{Results}
\begin{figure}[!t]
  \begin{center}
    \includegraphics[width=0.4\textwidth]{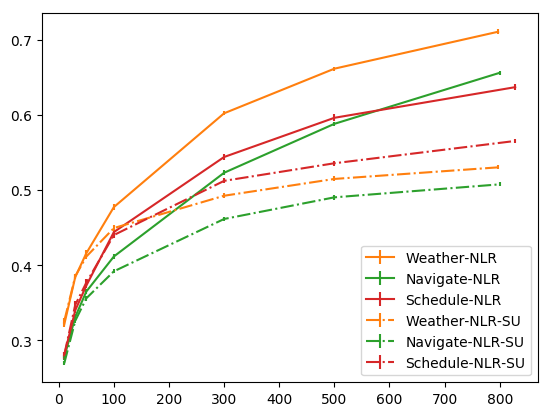}
  \end{center}
  \caption{Performance curves}
  \label{fig:perf}
\end{figure}

To evaluate our proposed model, we conducted ablation tests in Table.~\ref{tab:ablation} -- NLR: the proposed model, NLR-S: without {\em s-rules}, NLR-U: without {\em u-rules}, NLR-SU: no NLR inference. 
A more detailed description on the model parameters and training process can be found in Section~\ref{train}. 
The proposed model outperforms the baseline, NLR-SU, by about 6\% - 20\% in terms of {\em Recall@1}. Across all the domains, the performance consistently increases as more domain knowledge gets added. 
To investigate how much sample complexity is reduced by incorporating domain knowledge, we plot performance curves depending on the training data size in Figure~\ref{fig:perf}. NLR reaches about the same performance with only 100 - 300 dialogs that the baseline model, NLR-SU, achieves with full training data (around 800 dialogs). Finally, as the computation of cosine similarity in the NLR inferencer depends on the quality of sentence embeddings, we report performance with different sentence encoders in Table~\ref{tab:encoder} -- NC: Neurocon, ST: Skip-Thoughts~\cite{kiros2015skip}, WE: we train sentence RNNs with pre-trained word embeddings, SC: we train both sentence RNNs and word embeddings from scratch. The performance consistently increases as we use more pre-trained components. Although NC and ST both are pre-trained sentence encoders, the performance gap between between NC and ST indicates that, for conversational models, a sentence encoder trained on conversation data is better than one trained on plain text like books.

\subsection{Training Details}
\label{train}
For the sentence RNNs that we trained for the WE and SC models, we use a bidirectional LSTM-RNN with 100 hidden units for each direction. For the SC model, the word embedding weight matrix was initialized with the GloVe embeddings with 100 dimension~\cite{pennington2014glove}.
For the ST model, we obtained the Skip-Thoughts model~\cite{kiros2015skip} at the author's repository~\footnote{https://github.com/ryankiros/skip-thoughts} which was trained on the BookCorpus data. Specifically, we used the uni-skip model which uses a GRU-RNN~\cite{cho2014learning} encoder with 2,400 hidden units. 
For the context encoder, we use an LSTM-RNN with 200 hidden units. We initialized all the weights of the LSTM-RNN using the {\em Xavier} uniform distribution~\cite{glorot2010understanding}. For the matcher, we initialized $\lambda$ to 0.1. We use the Adam optimizer~\cite{kingma2014adam}, with gradients computed on mini-batches of size 1 and clipped with norm value 5. The learning rate was set to $1 \times 10^{-3}$ throughout the training and all the other hyperparameters were left as suggested in \cite{kingma2014adam}. We performed early stopping based on the performance of the evaluation data to avoid overfitting.

\newcolumntype{C}[1]{>{\centering\arraybackslash}p{#1}}
\begin{table}[t!]
\centering
\small
\begin{tabular}{|c|c|c|c|c|}
\hline
Domain	& NLR	& NLR-S	& NLR-U	& NLR-SU \\ \hline
\texttt{Weather}	& \bf{71.96}	& 61.62	& 67.90	& 52.77  \\ \hline
\texttt{Navigate}	& \bf{62.69}	& 57.80	& 59.94	& 48.62 \\ \hline
\texttt{Schedule}	& \bf{66.17}	& 63.68	& 61.69	& 58.21 \\ \hline
\end{tabular}
\caption{Ablation test results in {\em Recall@1}.}
\label{tab:ablation}
\end{table}

\begin{table}[t!]
\centering
\small
\begin{tabular}{|c|c|c|c|c|}
\hline
Domain	& NC	& ST	& WE	& SC \\ \hline
\texttt{Weather}	& \bf{71.96}	& 52.77		& 45.76		& 42.07  \\ \hline
\texttt{Navigate}	& \bf{62.69}	& 51.99		& 41.90		& 35.19 \\ \hline
\texttt{Schedule}	& \bf{66.17}	& 46.77		& 46.77		& 38.31 \\ \hline
\end{tabular}
\caption{Experimental results with different encoders.}
\label{tab:encoder}
\end{table}

\section{Conclusion}
\label{sec:conclusion}
We have presented a novel approach to improve the data-intensiveness problem of neural conversational models. We tackled the problem with the NLR inferencer that allows one to transfer domain knowledge in the form of natural language rules.
We tested our method with multiple dialog datasets. The improved performance across all domains demonstrates the efficacy of our proposed method.

\bibliographystyle{IEEEbib}
\bibliography{acl2018}

\end{document}